\def\year{2022}\relax
\documentclass[letterpaper]{article} 
\usepackage{aaai22}  
\usepackage{times}  
\usepackage{helvet}  
\usepackage{courier}  
\usepackage[hyphens]{url}  
\usepackage{graphicx} 
\usepackage{natbib}  
\usepackage{caption} 
\DeclareCaptionStyle{ruled}{labelfont=normalfont,labelsep=colon,strut=off} 
\usepackage{algorithm}
\usepackage{algorithmic}

\usepackage{newfloat}
\usepackage{listings}
\floatstyle{ruled}
\newfloat{listing}{tb}{lst}{}
\floatname{listing}{Listing}
\usepackage[colorlinks=true]{hyperref} 

\usepackage{enumerate}
\usepackage{enumitem}
\usepackage{bm}
\usepackage{amsmath}
\usepackage{amssymb}
\usepackage{amsfonts}
\usepackage{multirow}
\usepackage{tabularx}
\usepackage{mathrsfs}
\usepackage{amsthm}
\usepackage{booktabs}

\newtheorem{proposition}{Proposition}

\newcommand{\ie}{\emph{i.e.},~}
\newcommand{\eg}{\emph{e.g.}~}
\newcommand{\wrt}{\emph{w.r.t.}~}

\newcommand{\xiaok}[1]{\left(#1\right)}
\newcommand{\zhongk}[1]{\left[#1\right]}
\newcommand{\dak}[1]{\left\{#1\right\}}

\newcommand{\shuk}[1]{\left\lVert#1\right\rVert}

\newcommand{\argmax}[1]{{\mathop{\arg\mathrm{max}}_{#1}\,}}

\newcommand{\pfrac}[2]{\frac{\partial #1}{\partial #2}}

\newcommand{\biset}[1]{\{0,1\}^{#1}}

\newcommand{\opseq}[3]{{#1_1 #3 #1_2 #3 \cdots #3 #1_{#2}}}
\newcommand{\seq}[2]{\opseq{#1}{#2}{,}}

\newcommand{\bmb}{\bm{b}}
\newcommand{\bmc}{\bm{c}}

\newcommand{\bmp}{\bm{p}}

\newcommand{\bmx}{\bm{x}}

\newcommand{\bmz}{\bm{z}}

\newcommand{\bmtheta}{\bm{\theta}}

\newcommand{\bmC}{\bm{C}}

\newcommand{\bmXi}{\bm{\Xi}}

\newcommand{\calC}{\mathcal{C}}
\newcommand{\calD}{\mathcal{D}}

\newcommand{\calL}{\mathcal{L}}
\newcommand{\calM}{\mathcal{M}}
\newcommand{\calN}{\mathcal{N}}

\newcommand{\calQ}{\mathcal{Q}}
\newcommand{\calR}{\mathcal{R}}

\newcommand{\calT}{\mathcal{T}}

\newcommand{\calX}{\mathcal{X}}

\newcommand{\bbE}{\mathbb{E}}

\newcommand{\bbR}{\mathbb{R}}

\title{Contrastive Quantization with Code Memory for Unsupervised Image Retrieval}
\author{
    Jinpeng Wang\textsuperscript{\rm 1,2,4}, Ziyun Zeng\textsuperscript{\rm 1,4}, Bin Chen\textsuperscript{\rm 2}\thanks{Corresponding author.}, Tao Dai\textsuperscript{\rm 3}, Shu-Tao Xia\textsuperscript{\rm 1,4}
}
\affiliations{
    \textsuperscript{\rm 1}Tsinghua Shenzhen International Graduate School, Tsinghua University\\
    \textsuperscript{\rm 2}Harbin Institute of Technology, Shenzhen\\ 
    \textsuperscript{\rm 3}Shenzhen University\\
    \textsuperscript{\rm 4}Research Center of Artificial Intelligence, Peng Cheng Laboratory \\
    \{wjp20, zengzy21\}@mails.tsinghua.edu.cn,
    chenbin2021@hit.edu.cn,
    daitao.edu@gmail.com,
    xiast@sz.tsinghua.edu.cn
}

\usepackage{bibentry}

\begin{document}
\maketitle
\begin{abstract}
The high efficiency in computation and storage makes hashing (including binary hashing and quantization) a common strategy in large-scale retrieval systems. 
To alleviate the reliance on expensive annotations, unsupervised deep hashing becomes an important research problem. 
This paper provides a novel solution to \emph{unsupervised deep quantization}, namely \textbf{Co}ntrastive \textbf{Q}uantization with Code \textbf{Me}mory (\textbf{MeCoQ}). 
Different from existing reconstruction-based strategies, we learn unsupervised binary descriptors by contrastive learning, which can better capture discriminative visual semantics. 
Besides, we uncover that \emph{codeword diversity regularization} is critical to prevent contrastive learning-based quantization from model degeneration. 
Moreover, we introduce a novel \emph{quantization code memory module} that boosts contrastive learning with lower feature drift than conventional feature memories. 
Extensive experiments on benchmark datasets show that MeCoQ outperforms state-of-the-art methods.
Code and configurations are publicly available at \url{https://github.com/gimpong/AAAI22-MeCoQ}.
\end{abstract}

\section{Introduction}
Hashing~\cite{l2h2} plays a key role in Approximate Nearest Neighbor (ANN) search and has been widely applied in large-scale systems to improve search efficiency. 
There are two technical branches in hashing, namely binary hashing and quantization.
Binary hashing methods~\cite{lsh,sphh} transform data into the Hamming space such that distances are measured quickly with bitwise operations. 
Quantization methods~\cite{pq} divide real data space into disjoint cells. Then the data points in each cell are approximately represented as the centroid. Since the inter-centroid distances can be pre-computed as a lookup table, quantization methods can efficiently calculate pairwise distance. 

With the progress in deep learning, the past few years have seen many deep hashing methods~\cite{csq,dsh} with impressive performance. 
Unfortunately, annotating tons of data in real-world applications is expensive, making it hard to apply these supervised methods. 
Recent research interests have arisen in unsupervised deep hashing to address this issue, but existing works are not satisfactory enough. 
On the one hand, most existing studies in unsupervised deep hashing focus on preserving the information from continuous features. They mostly use quantization loss~\cite{deepbit,deepquan} and similarity reconstruction loss~\cite{sadh} as the learning objectives, resulting in heavy reliance on the quality of extracted features from pre-trained backbones~\cite{alexnet,vgg,resnet}. If an adopted backbone generalizes poorly in the target domain, the unsatisfactory features will degrade the output binary codes fundamentally. 
On the other hand, \citet{deepbit,uth} introduced rotation invariance of images to learn deep hashing, but weak negative samples and ineffective training schemes led to inferior performance.

This paper focuses on \emph{unsupervised deep quantization}. 
To make better use of unlabeled training data, we perform Contrastive Learning (CL)~\cite{hadsell2006dimensionality} that learns representations by mining visual-semantic invariance from inputs~\cite{simclr,moco}. 
CL has become a promising direction toward deep unsupervised representation, but CL-based deep quantization remains non-trivial. 
Specifically, we find three challenges in this task:
(\textbf{i}) \emph{Sampling bias}. Without label supervision, a randomly sampled batch may contain positive samples that are falsely taken as negatives. 
(\textbf{ii}) \emph{Model degradation}. We observe that quantization codewords of the same codebook tend to get closer during CL, which gradually degrades the representation ability and harms the model.
(\textbf{iii}) \emph{The conflict between effect and efficiency in training}. 
CL benefits from a large batch size that ensures enough negative samples, while a single GPU can afford a limited batch size. 
Training by CL often requires multi-GPU synchronization, which is complex in engineering and less efficient. 
To improve the efficiency, some recent studies~\cite{instdisc,pirl} enable small-batch CL by caching embeddings in a memory bank and reusing them as negatives in later iterations. 
However, as the encoder updates, the cached embeddings will expire and affect the effect of CL.

To tackle the problems, we propose \textbf{Co}ntrastive \textbf{Q}uantiza-tion with Code \textbf{Me}mory (\textbf{MeCoQ}) that combines memory-based CL and deep quantization in a \emph{mutually beneficial} framework. 
Specifically, 
(\textbf{i}) \emph{MeCoQ is bias-aware}. 
We adopt a debiased framework~\cite{dcl} that can correct the sampling bias in CL. 
(\textbf{ii}) \emph{MeCoQ avoids degeneration}. 
We find that codeword diversity is critical to prevent the CL-based quantization model from degeneration. Hence, we design a codeword regularization to reinforce MeCoQ.
(\textbf{iii}) \emph{MeCoQ boosts CL effectively and efficiently}. 
We propose a novel memory bank for quantization codes 
that shows lower feature drift~\cite{xbm} than existing feature memories. 
Thus, it can retain cached negatives valid for a longer period and enhance the effect of CL, without heavy computations from the momentum encoder~\cite{moco}. 

Our contributions can be summarized as follows.
\setlist{nolistsep}
\begin{itemize}
	\item[$\bullet$] We provide a novel solution to unsupervised deep quantization, which combines contrastive learning and deep quantization in a mutually beneficial framework.
	\item[$\bullet$] We show that codeword diversity is critical to prevent contrastive deep quantization from model degeneration.
	\item[$\bullet$] We propose a quantization code memory to enhance the effect of memory-based contrastive learning.
	\item[$\bullet$] Extensive experiments on public benchmarks show that MeCoQ outperforms state-of-the-art methods.
\end{itemize}

\section{Related Work}
\subsubsection{Unsupervised Hashing}
Traditional unsupervised hashing includes binary hashing~\cite{lsh,speh,sh,sphh,itq} and quantization~\cite{pq,opq,cq,unq}. 
Limited by the hand-designed representations~\cite{gist,sift}, these methods reach suboptimal performance.

Deep hashing methods with Convolutional Neural Networks (CNNs) \cite{alexnet,vgg,resnet} usually perform better than non-deep hashing methods. 
Existing deep hashing methods can be categorized into generative~\cite{sgh,dbdmq,bingan,bgan,hashgan,dvb,tbh,bihalf,cibhash} or discriminative~\cite{deepbit,uth,greedyhash,deepquan,ssdh,distillhash,mls3rduh} series. 
Most of them impose various constraints (\ie loss or regularization terms) such as pointwise constraints:
(\textbf{i}) quantization error~\cite{dbdmq,deepquan}, 
(\textbf{ii}) even bit distribution~\cite{bingan,dvb}, 
(\textbf{iii}) bit irrelevance~\cite{hashgan}, 
(\textbf{iv}) maximizing mutual information between features and codes~\cite{bihalf,cibhash}; 
and pairwise constraints: 
(\textbf{v}) preserving similarity among continuous feature vectors~\cite{greedyhash,ssdh,distillhash,mls3rduh}. 
They merely explore statistical characteristics of hash codes or focus on preserving the semantic information from continuous features, which leads to heavy dependence on high-quality pre-trained features. 
Due to limited generalization ability, the adopted CNNs may extract unsatisfactory features for images from new domains, which harms generated hash codes fundamentally. 
On the other hand, DeepBit~\cite{deepbit} and UTH~\cite{uth} introduce rotation invariance of images to improve deep hashing. 
Unfortunately, rotation itself is not enough to construct informative negative samples and the training schemes of DeepBit and UTH are less effective, which leads to inferior performances. 
Most recently, \citet{cibhash} combined contrastive learning with deep binary hashing. 
By taking effective data augmentations and engaging more negative samples, it shows promising results. 
Different from them, we explore the combination of contrastive learning and deep quantization that is more challenging. 
We propose a codeword diversity regularization to prevent model degeneration. 
Besides, we adopt a debiasing mechanism and propose a quantization code memory to enhance contrastive learning, yielding better results.


\subsubsection{Contrastive Learning}
Contrastive Learning (CL) \cite{hadsell2006dimensionality} based representation learning has drawn increasing attention. \citet{instdisc} proposed an instance discrimination method that combines a non-parametric classifier (\ie a memory bank) with a cross-entropy loss (\emph{aka} InfoNCE \cite{cpc} or contrastive loss \cite{moco}). Positive samples from the same image are pulled closer and negative samples from other images are pushed apart. 
The subsequent instance-wise CL methods focus on designing \emph{end-to-end}~\cite{simclr}, \emph{memory bank}-based~\cite{pirl}, or \emph{momentum encoder}-based~\cite{moco} architectures. 
Besides, cluster-wise contrastive methods~\cite{swav,pcl} integrate the clustering objective into CL, which shows promising results. 
Moreover, there are some works addressing sampling bias~\cite{dcl} or exploring effective negative sampling~\cite{hcl} in CL, which improve the training effect. 
We uncover that codeword diversity is critical to enable CL in deep quantization. 
Besides, we propose a novel memory that stores quantization codes. 
Interestingly, without needing a momentum encoder, it can show lower feature drift~\cite{xbm} than existing feature memories~\cite{instdisc,pirl} and thus boosts CL effectively. 

\begin{figure*}[t]
  \centering
  \includegraphics[width=0.95\textwidth]{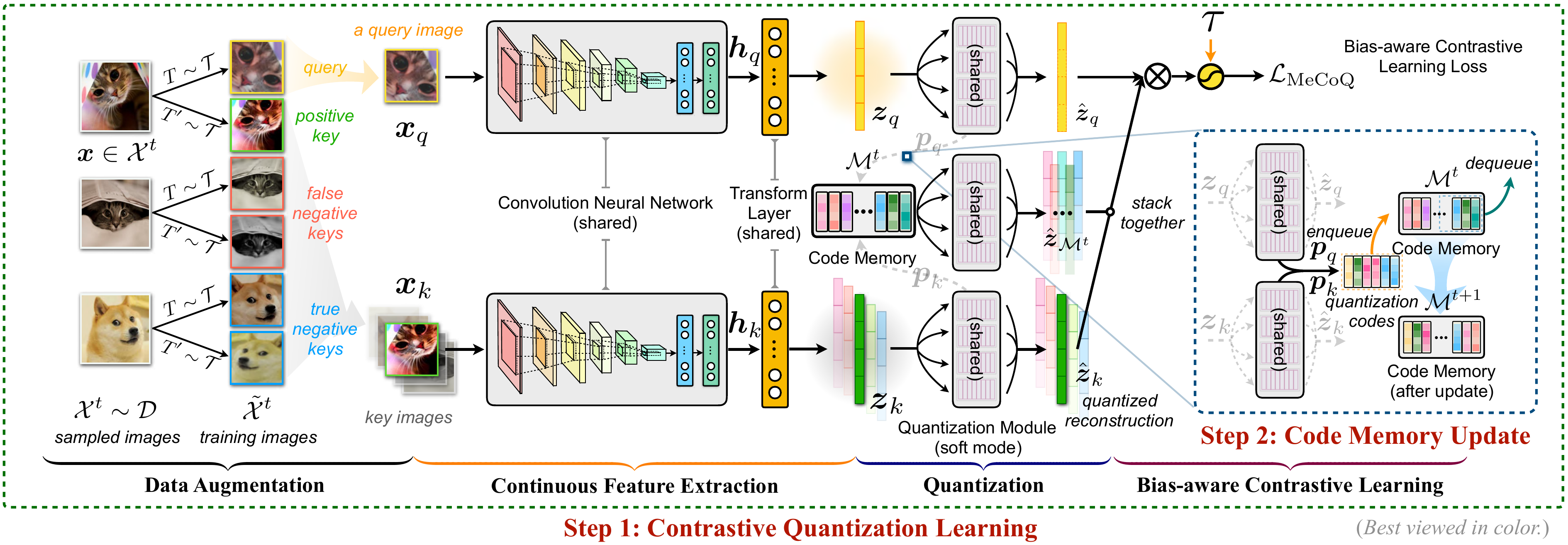}
  \caption{The framework of MeCoQ. After augmentation, we set one view for an image as the training \emph{query} $\bmx_q$ and leave the other view $\bmx_k$ along with the views of other images as the \emph{keys} in contrastive learning. Then, we extract the embeddings for these images and forward them to the quantization module to get quantized reconstructions. Embeddings reconstructed from the code memory bank serve as additional negative keys that boost contrastive learning. Next, we maximize the similarity between the query and the positive key (\ie the other view from the same image) and minimize the similarities of negative query-key pairs. Finally, we update the code memory bank with the quantization codes of the views of current image batch.}
\label{fig:arc}
\end{figure*}



\section{Modeling Framework}

\subsection{Problem Formulation and Model Overview}
Given an unlabeled training set
$\calD$ of $N_\calD$ images where each image $\bmx$ can be flattened as a $P$-dimensional vector, 
the goal of unsupervised deep quantization is to learn a neural quantizer $Q:\bbR^P\mapsto\biset{B}$ that encodes images as $B$-bit semantic quantization codes (\emph{aka} the binary representations) for efficient image retrieval. 
To this end, we propose \textbf{Co}ntrastive \textbf{Q}uantization with Code \textbf{Me}mory (\textbf{MeCoQ}) in an end-to-end deep learning architecture. 
As shown in Figure~\ref{fig:arc}, MeCoQ consists of:
	(\textbf{i})
	Two operators sampled from the same data augmentation family ($T\sim\calT$ and $T'\sim\calT$), which are applied to each training image to obtain two correlated views ($\bmx_q$ and $\bmx_k\in\bbR^P$). 
	(\textbf{ii})
	A deep embedding module $h$ combined with a standard CNN~\cite{vgg} and a transform layer, which produces a continuous embedding $\bmz\in\bbR^D$ for an input view $\bmx\in\bbR^P$. 
	(\textbf{iii})
	A trainable quantization module that produces a \emph{soft} quantization code vector $\bmp$ and the reconstruction $\hat{\bmz}\in\bbR^D$ for $\bmz\in\bbR^D$. In inference, it directly outputs the \emph{hard} quantization code vector $\bmb\in\biset{B}$ for image $\bmx$.
	(\textbf{iv})
	A code memory bank $\calM$ to cache the quantization code vectors of images, which serves as an additional source of negative training keys and is not involved in inference.

\subsection{Debiased Contrastive Learning for Quantization}
\subsubsection{Trainable Quantization}
It is hard to integrate traditional quantization~\cite{pq} into the deep learning framework because the codeword assignment step is clustering-based and can not be trained by back-propagation. 
To enable end-to-end learning in MeCoQ, we apply a trainable quantization scheme. 
Denote the quantization codebooks as $\calC=\bmC^1\times\bmC^2\times\cdots\times\bmC^M$, where the $m$-th codebook $\bmC^m\in\bbR^{K\times d}$ consists of $K$ codewords $\seq{\bmc^m}{K}\in\bbR^d$. 
Assume that $\bmz\in\bbR^D$ can be divided into $M$ equal-length $d$-dimensional segments, \ie $\bmz\in\bbR^D\equiv[\bmz^1,\cdots,\bmz^M]$, $\bmz^m\in\bbR^d$, $d=D/M$, $1\le m\le M$.
Given a vector, each codebook is used to quantize one segment respectively.
In the $m$-th $d$-dimensional subspace, the segment and codewords are first normalized:
\begin{equation}
    \bmz^m\leftarrow\bmz^m/\shuk{\bmz^m}_2,\ \bmc_i^m\leftarrow\bmc_i^m/\shuk{\bmc_i^m}_2.
\end{equation}
Then each segment is quantized with codebook attention by
\begin{equation}\label{equ:code_attention}
    \hat{\bmz}^m=\text{Attention}(\bmz^m,\bmC^m,\bmC^m)
    =\sum_{i=1}^Kp_i^m\bmc_i^m,
\end{equation}
where attention score $p_i^m$ is computed with the $\alpha$-softmax:
\begin{equation}\label{equ:alpha_softmax}
    p^m_i
    =\text{softmax}_\alpha\xiaok{{{\bmz^m}^\top\bmc_i^m}}
    =\frac{\exp\xiaok{\alpha\cdot{\bmz^m}^\top\bmc^m_i}}{\sum_{j=1}^K\exp\xiaok{\alpha\cdot{\bmz^m}^\top\bmc^m_j}}.
\end{equation}
The $\alpha$-softmax is a differentiable alternative to the argmax that relaxes the discrete optimization of hard codeword assignment to a trainable form. 
Finally, we get \emph{soft} quantization code 
and the \emph{soft} quantized reconstruction of $\bmz$ as
\begin{gather}
    \bmp=\text{concatenate}\xiaok{\bmp^1,\bmp^2,\cdots,\bmp^M}\in\bbR^{KM},\\
    \label{equ:concat}\hat{\bmz}= \text{concatenate}\xiaok{\hat{\bmz}^1,\hat{\bmz}^2,\cdots,\hat{\bmz}^M}.
\end{gather}

\subsubsection{Debiased Contrastive Learning}
We conduct Contrastive Learning (CL) based on the soft quantized reconstruction vectors. 
Because negative keys for a training query are randomly sampled from unlabeled training set, there are unavoidably some \emph{false}-negative keys that harm CL. 
To tackle this problem, we adopt a bias-aware framework~\cite{dcl}. The debiased CL loss is defined as 
\begin{equation}\label{equ:dcl}
\calL_\text{DCL}=-\sum_{q=1}^{2N}\log\frac{\exp(\frac{s_{q,k^+}}{\tau})}
{\exp(\frac{s_{q,k^+}}{\tau})+\calN_\text{In-Batch}},
\end{equation}
where $N$ is the batch size, $q$ and $k^+$ denote the indices of the training query and the positive key in the augmented batch, $\tau$ is the temperature hyper-parameter. 
The similarity between a query $q$ and a key $k$ is defined as $s_{q,k}\triangleq\hat{\bmz}_q^\top\hat{\bmz}_k$.
The debiased in-batch negative term in Eq.(\ref{equ:dcl}) is defined as
\begin{equation}
\label{equ:inbatch}
    \calN_\text{In-Batch} \triangleq \sum_{\substack{k^-=1,\\k^-\notin\{q, k^+\}}}^{2N}\zhongk{\frac{\exp(\frac{s_{q,k^-}}{\tau})}{1-\rho^+}-\frac{\rho^+\cdot\exp(\frac{s_{q,k^+}}{\tau})}{1-\rho^+}}, 
\end{equation}
where $k^-$ denotes the indices of negative key and $\rho^+$ is the positive prior for bias correction. 

\begin{figure}[t]
  \centering
  \includegraphics[width=\columnwidth]{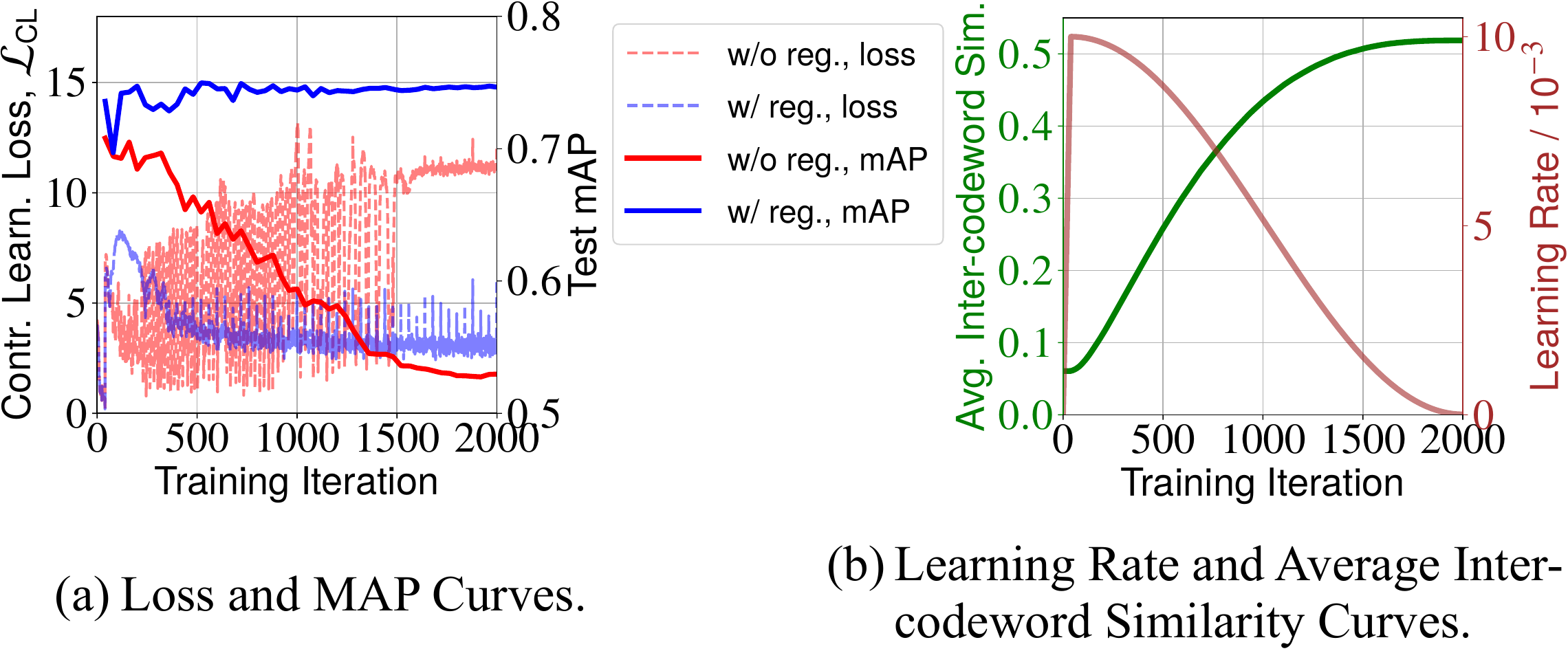}
  \caption{MeCoQ degenerates during contrastive learning because codewords in the same codebook are getting closer. The codeword diversity regularization can avoid it.}
\label{fig:degrade}
\end{figure}

\subsection{Regularization to Avoid Model Degeneration}\label{sec:codebook_reg}
\subsubsection{Model Degeneration}
We find that deep quantization is prone to degenerate as CL goes on. 
To show the isuue, we reduce irrelevant factors\footnote{We have done pretest about the effect of debiasing mechanism to the issue. 
As the results showed that it does not change the phenomenon, we exclude it for a concise interpretation in this section.} and train a simplified 32-bit MeCoQ by vanilla CL on the Flickr25K dataset. 
As shown in Figure~\ref{fig:degrade}(a), the model performance (solid red line) declines while the loss (dash red line) rises with fluctuation. 
It seems that the success of CL for continuous representation doesn't generalize to quantization directly. 
Considering the difference in models, we investigate the behavior of quantization codebooks during CL. 
Intuitively, we observe the changes of average inter-codeword similarity, namely, 
\begin{equation}
    \label{equ:codeword_reg}
    \Omega_\calC = \frac1{MK^2}\sum_{m=1}^M\sum_{i=1}^K\sum_{j=1}^K{\bmc^m_i}^\top{\bmc^m_j}. 
\end{equation} 
Figure~\ref{fig:degrade}(b) shows a monotonic increase of $\Omega_\calC$, which slows down as the learning rate decreases. 
It suggests that the optimization leads to a degenerated solution. 

Here we discuss what may cause this phenomenon. Recall that vanilla CL loss \wrt a training query $\bmx_q$ is
\begin{equation}
\calL_\text{CL}(\bmx_q)=-\log\frac{\exp(\frac{s_{q,k^+}}{\tau})}
{\exp(\frac{s_{q,k^+}}{\tau})+\sum_{\substack{k^-=1,\\k^-\notin\{q, k^+\}}}^{2N}\exp(\frac{s_{q,k^-}}{\tau})}.
\end{equation}
\begin{proposition}
Suppose that $\bmx_{k^+}$ is the positive training key and $\bmx_{k^-}$ is a negative key to $\bmx_q$. $\hat{\bmz}_{k^+}$ and $\hat{\bmz}_{k^-}$ are the reconstructed embeddings \wrt $\bmx_{k^+}$ and $\bmx_{k^-}$, then we have
\begin{equation} 
    \shuk{\pfrac{\calL_\text{\emph{CL}}(\bmx_q)}{\hat{\bmz}_{k^+}}}>\shuk{\pfrac{\calL_\text{\emph{CL}}(\bmx_q)}{\hat{\bmz}_{k^-}}}>0.
\end{equation}
\end{proposition}

We prove Proposition 1 in Appendix. 
It suggests that the scale of loss gradient \wrt the positive key is greater than that \wrt any negative key. Engaging more negatives leads to a larger gap between such scales. 
Besides, to reduce the deviation between soft quantization in training and hard quantization in inference, we take a relatively large $\alpha$ (10 by default) in the codeword attention (Eq.(\ref{equ:code_attention})). 
It makes each soft reconstruction and assigned codeword approximate at the forward step. In the back-propagation, their gradients are also approximate. Thus, it can hold when replacing $\hat{\bmz}_{k^+}$ and $\hat{\bmz}_{k^-}$ in Proposition 1 with assigned codewords.
If the query and the positive key are assigned to different codewords, there will be a large gradient to pull these codewords closer. 
We find it irreversible without explicit control, because of the insufficient frequency of the same assignment for negative pairs along with subtle gradients to push the same of codewords away. 
As a result, the representation ability of codebooks degrades and the model degenerates.

\subsubsection{Codeword Diversity Regularization}
To avoid degeneration, we regularize the optimization by imposing $\Omega_\calC\le\epsilon$, where $\epsilon$ is a fixed bound. 
In practice, we set it as a loss term that encourages codeword diversity and guides deep quantization model to pay more attention to proper codeword assignment rather than violently moving the codewords. 
The blue lines in Figure~\ref{fig:degrade}(a) show that the regularization effectively avoids the issue and also helps to calm the loss down.

\subsection{Quantization Code Memory to Boost Training}
Effective CL methods rely on sufficient negative samples to learn discriminative representations. 
Therefore, existing memory-based methods cache the image embeddings and serve them as the negatives in the later training. 
However, as the model keeps updating, the early cached embeddings expire and become noises to CL. 
To enhance the effect of memory, MoCo~\cite{moco} introduces a momentum encoder that mitigates the embedding aging issue at a higher computation cost. 
In contrast, we find an elegant solution that achieves a similar effect more efficiently.

\subsubsection{Feature Drift}
We follow~\citet{xbm} to investigate the embedding aging issue. 
We define the \emph{feature drift} of a neural embedding model $h$ by 
\begin{equation}
    \text{Drift}(\calX',t;\Delta t)\triangleq\frac1{|\calX'|}\sum_{\bmx\in\calX'}\shuk{h(\bmx;\bmtheta_h^t)-h(\bmx;\bmtheta_h^{t-\Delta t})}_2^2,
\end{equation}
where $\calX'$ is a given image set for estimation, $\bmtheta_h$ is the parameters of $h$, $t$ and $\Delta t$ denote the number and the interval of training iterations (\ie batches) respectively. 
We train a 32-bit CL-based quantization model
and compute its feature drift based on original embeddings, soft quantized reconstructions, and hard quantized reconstructions. 

\begin{figure}[t]
  \centering
  \includegraphics[width=\columnwidth]{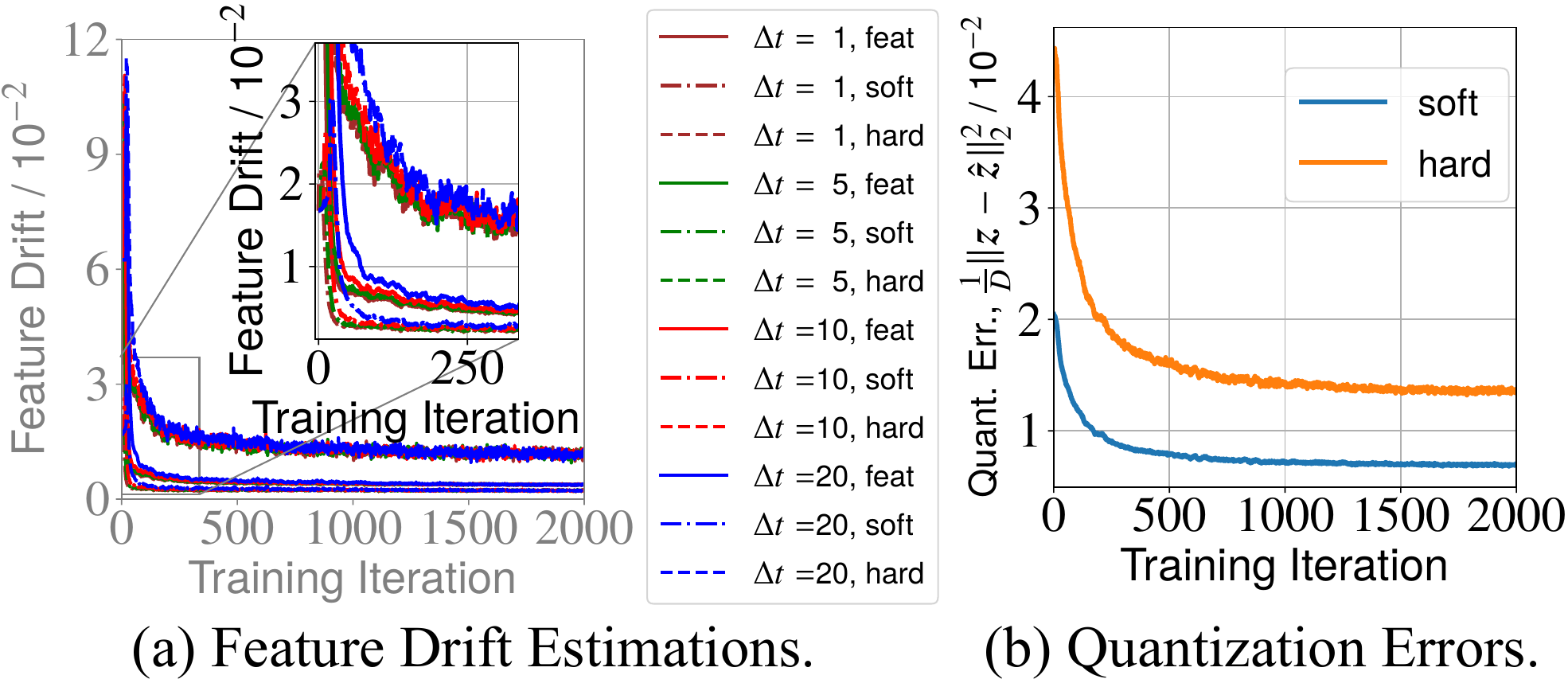}
  \caption{We estimate the feature drifts of original features, soft quantized features, and hard quantized features. The soft features show lower drift than the original features because moderate quantization error can compensate for the drift.}
\label{fig:memory}
\end{figure}

As shown in Figure~\ref{fig:memory}(a), the embeddings violently change at the early stage, after which they become relatively stable. 
Moreover, it is surprising that the \emph{soft} quantized embeddings show even lower feature drifts than the original embeddings. 
We realize that undesirable quantization error (illustrated in Figure~\ref{fig:memory}(b)) that used to be eliminated in quantization models \emph{partially}\footnote{Note that \emph{hard} quantization shows higher feature drift. It doesn't benefit from the offset because of large quantization error.} offsets the feature drift, keeping the cached information valid for a longer time. 

\subsubsection{Memory-augmented Training}
Base on the above facts, we start using a memory bank $\calM$ with $N_\calM$ slots to store \emph{soft quantization codes} after the warm-up stage. 
In each training iteration, we fetch the cached codes $\bmp_{\calM_1}, \bmp_{\calM_2}, \cdots, \bmp_{\calM_{N_\calM}}$ from the memory bank. 
Then, we forward them to the quantization module and respectively reconstruct the embeddings $\hat{\bmz}_{\calM_1}, \hat{\bmz}_{\calM_2},\cdots,\hat{\bmz}_{\calM_{N_\calM}}$ by Eq.(\ref{equ:code_attention}) and~(\ref{equ:concat}). 
Finally, we integrate these embeddings to Eq.(\ref{equ:dcl}) and formulate the memory-augmented loss as
\begin{equation}\label{equ:mecoq}
    \calL_\text{MeCoQ}=-\sum_{q=1}^{2N}\log\frac{\exp(\frac{s_{q,k^+}}{\tau})}
{\exp(\frac{s_{q,k^+}}{\tau})+\calN_\text{In-Batch}+\calN_\text{Memory}},
\end{equation}
where the added negative term about code memory is
\begin{equation}
\label{equ:memory}
    \calN_\text{Memory} \triangleq  \sum_{i=1}^{N_\calM}\zhongk{\frac{\exp(\frac{s_{q,\calM_i}}{\tau})}{1-\rho^+}-\frac{\rho^+\cdot\exp(\frac{s_{q,k^+}}{\tau})}{1-\rho^+}}.
\end{equation}

At the end of each training iteration, we update the memory as a queue, \ie the current batch is enqueued and an equal number of the oldest slots are removed. To simplify engineering, we set the queue size $N_\calM$ to a multiple of the batch size $N$ so that we update the memory bank by batches.

\subsection{Learning Algorithm}
The learning objective of MeCoQ is
\begin{equation}
    \min_{\bmtheta_h,\calC}\bbE\,\calL_\text{MeCoQ}+\beta\|\bmtheta_h\|_2^2 + \gamma\Omega_\calC,
\end{equation}
where $\calL_\text{MeCoQ}$ is formulated as Eq.(\ref{equ:mecoq}), $\Omega_\calC$ is defined as Eq.(\ref{equ:codeword_reg}). $\bmtheta_h$ denotes the network parameters of the embedding module $h$, $\beta$ and $\gamma$ are the trade-off hyper-parameters. 
The training process is quite efficient as we formulate the whole problem in a deep learning framework. Many off-the-shelf optimizers can be applied within a few code lines.


\subsection{Encoding and Retrieval}
In inference, we encode the database with hard quantization:
\begin{gather}
    i^m_{db}=\argmax{1\le i\le K}{{\bmz^m_{db}}^\top\bmc_i^m},\qquad
    \hat{\bmz}^m_{db}=\bmc_{i^m_{db}}^m.
\end{gather}
We can aggregate the indices $\zhongk{i^1_{db},i^2_{db},\cdots,i^M_{db}}$ and convert it into a code vector $\bmb_{db}$ for the database image $\bmx_{db}$.
Given a retrieval query image $\bmx_q$, we extract its deep embedding and cut the vector into $M$ equal-length segments, \ie $\bmz_q=[\bmz_q^1; \bmz_q^2; \cdots; \bmz_q^M]$, $\bmz_q^m\in\bbR^d$. We adopt Asymmetric Quantized Similarity (AQS)~\cite{pq} as the metric, which computes the similarity between $\bmz_q$ and the reconstruction of a database point $\hat{\bmz}_{db}$ by
\begin{equation}
    \text{AQS}(\bmx_q, \bmx_{db})=\sum_{m=1}^M\frac{{\bmz_q^m}^\top\hat{\bmz}_{db}^m}{\shuk{\bmz_q^m}_2}=\sum_{m=1}^M\frac{{\bmz_q^m}^\top\bmc_{i^m_{db}}^m}{\shuk{\bmz_q^m}_2}.
\end{equation}
We can set up a query-specific lookup table $\bmXi_q\in\bbR^{M\times K}$ for each $\bmx_q$, which stores the pre-computed similarities between the segments of $\bmx_q$ and all codewords. 
Specifically, $\Xi^m_{q,i^m_{db}}={\bmz_q^m}^\top\bmc_{i^m_{db}}^m/\|\bmz^m_q\|_2$. 
Hence, the AQS can be efficiently computed by summing chosen items from the lookup table according to the quantization code, \ie
\begin{equation}
    \text{AQS}(\bmx_q, \bmx_{db})=\sum_{m=1}^M\Xi^m_{q,i^m_{db}},
\end{equation}
where $i^m_{db}$ is the index of codeword in the $m$-th codebook. 

\begin{table*}[t]
    \centering
    \caption{Mean Average Precision (MAP, \%) results for different number of bits on Flickr25K, CIFAR-10 (I and II) and NUS-WIDE datasets. `D', `Q' and `BH' indicate `Deep', `Quantization' and `Binary Hashing' for short in `Type' column.}
    \resizebox{\linewidth}{!}{
    \setlength{\tabcolsep}{.4em}{
    \begin{tabular}{lcclccclccclccclccc}
    \toprule
       & \multicolumn{2}{c}{Dataset $\rightarrow$} &  & \multicolumn{3}{c}{Flickr25K} &  & \multicolumn{3}{c}{CIFAR-10 (I)} &  & \multicolumn{3}{c}{CIFAR-10 (II)} & \multicolumn{1}{c}{} & \multicolumn{3}{c}{NUS-WIDE} \\
      \cmidrule(l){2-3} \cmidrule(l){5-7} \cmidrule(l){9-11} \cmidrule(l){13-15} \cmidrule(l){17-19}
      Method $\downarrow$ & Venue $\downarrow$ & Type $\downarrow$ &  & 16 bits & 32 bits & 64 bits & & 16 bits & 32 bits & 64 bits & & 16 bits & 32 bits & 64 bits & & 16 bits & 32 bits & 64 bits \\
    \midrule
      LSH+VGG & STOC'02 & BH &  & 56.11 & 57.08 & 59.26 & & 14.38 & 15.86 & 18.09 & & 12.55 & 13.76 & 15.07 & & 38.52 & 41.43 & 43.89 \\
      SpeH+VGG
      & NeurIPS'08 & BH &  & 59.77 & 61.36 & 64.08 & & 27.09 & 29.44 & 32.65 & & 27.20 & 28.50 & 30.00 & & 51.70 & 51.10 & 51.00 \\
      SH+VGG & IJAR'09 & BH &  & 60.02 & 63.30 & 64.17 & & 28.28 & 28.86 & 28.51 & & 24.68 & 25.34 & 27.16 & & 46.85 & 53.63 & 56.28 
      \\
      SphH+VGG & CVPR'12 & BH &  & 61.32 & 62.47 & 64.49 & & 26.90 & 31.75 & 35.25
      & & 25.40 & 29.10 & 33.30 & & 49.50 & 55.80 & 58.20 \\
      ITQ+VGG & TPAMI'12 & BH &  & 63.30 & 65.92 & 68.86 & & 34.41 & 35.41 & 38.82 & & 30.50 & 32.50 & 34.90 & & 62.70 & 64.50 & 66.40 \\
      PQ+VGG & TPAMI'10 & Q &  & 62.75 & 66.63 & 69.40 & & 27.14 & 33.30 & 37.67 & & 28.16 & 30.24 & 30.61 & & 65.39 & 67.41 & 68.56 \\
      OPQ+VGG & CVPR'13 & Q &  & 63.27 & 68.01 & 69.86 & & 27.29 & 35.17 & 38.48 & & 32.17 & 33.50 & 34.46 & & 65.74 & 68.38 & 69.12 \\
    \midrule
      DeepBit & CVPR'16 & DBH &  & 62.04 & 66.54 & 68.34 & & 19.43 & 24.86 & 27.73 & & 20.60 & 28.23 & 31.30 & & 39.20 & 40.30 & 42.90 \\
      UTH & ACMMMW'17 & DBH &  & - & - & - & & - & - & - & & - & - & - & & 45.00 & 49.50 & 54.90 \\
      SAH & CVPR'17 & DBH &  & - & - & - & & 41.75 & 45.56 & 47.36 & & - & - & - & & - & - & - \\
      SGH & ICML'17 & DBH &  & 72.10 & 72.84 & 72.83 & & 34.51 & 37.04 & 38.93 & & 43.50 & 43.70 & 43.30 & & 59.30 & 59.00 & 60.70 \\
      HashGAN & CVPR'18 & DBH &  & 72.11 & 73.25 & 75.46 & & 44.70 & 46.30 & 48.10 & & 42.81 & 47.54 & 47.29 & & 68.44 & 70.56 & 71.71 \\
      GreedyHash & NeurIPS'18 & DBH &  & 69.91 & 70.85 & 73.03 & & 44.80 & 47.20 & 50.10 & & 45.76 & 48.26 & 53.34 & & 63.30 & 69.10 & 73.10 \\
      BinGAN & NeurIPS'18 & DBH &  & - & - & - & & - & - & - & & 47.60 & 51.20 & 52.00 & & 65.40 & 70.90 & 71.30 \\
      BGAN & AAAI'18 & DBH &  & - & - & - & & - & - & - & & 52.50 & 53.10 & 56.20 & & 68.40 & 71.40 & 73.00 \\
      SSDH & IJCAI'18 & DBH &  & 75.65 & 77.10 & 76.68 & & 36.16 & 40.37 & 44.00 & & 33.30 & 38.29 & 40.81 & & 58.00 & 59.30 & 61.00 \\
      DVB & IJCV'19 & DBH &  & - & - & - & & - & - & - & & 40.30 & 42.20 & 44.60 & & 60.40 & 63.20 & 66.50 \\
      DistillHash & CVPR'19 & DBH &  & - & - & - & & - & - & - & & - & - & - & & 62.70 & 65.60 & 67.10 \\
      TBH & CVPR'20 & DBH &  & 74.38 & 76.14 & 77.87 & & 54.68 & 58.63 & 62.47 & & 53.20 & 57.30 & 57.80 & & 71.70 & 72.50 & 73.50 \\
      MLS$^3$RDUH & IJCAI'20 & DBH &  & - & - & - & & - & - & - & & - & - & - & & 71.30 & 72.70 & 75.00 \\
      Bi-half Net & AAAI'21 & DBH &  & 76.07 & 77.93 & 78.62 & & 56.10 & 57.60 & 59.50 & & 49.97 & 52.04 & 55.35 & & 76.86 & 78.31 & 79.94 \\
      CIBHash & IJCAI'21 & DBH &  & 77.21 & 78.43 & 79.50 & & 59.39 & 63.67 & 65.16 & & 59.00 & 62.20 & 64.10 & & 79.00 & 80.70 & 81.50 \\
      DBD-MQ & CVPR'17 & DQ &  & - & - & - & & 21.53 & 26.50 & 31.85 & & - & - & - & & - & - & - \\
      DeepQuan & IJCAI'18 & DQ &  & - & - & - & & 39.95 & 41.25 & 43.26 & & - & - & - & & - & - & - \\
    \midrule
      MeCoQ (Ours) & AAAI'22 & DQ &  & \textbf{81.31} & \textbf{81.71} & \textbf{82.68} && \textbf{68.20} & \textbf{69.74} & \textbf{71.06} && \textbf{62.88} & \textbf{64.09} & \textbf{65.07} && \textbf{80.18} & \textbf{82.16} & \textbf{83.24} \\
    \bottomrule
    \end{tabular}}}
    \label{tab:map1}
\end{table*}

\section{Experiments}

\subsection{Setup}
\subsubsection{Datasets}
(\textbf{i}) \textbf{Flickr25K}~\cite{mirflickr} contains
25k images from 24 categories.
We follow \citet{bihalf} to randomly pick 2,000 images as the testing queries, while another 5,000 images are randomly selected from the rest of the images as the training set.
(\textbf{ii}) \textbf{CIFAR-10}~\cite{cifar} contains 60k images from 10 categories. 
We consider two typical experiment protocols. 
\emph{CIFAR-10 (I)}:
We follow \citet{bihalf} to use 1k images per class (totally 10k images) as the test query set, and the remaining 50k images are used for training.
\emph{CIFAR-10 (II)}:
Following \citet{cibhash} we randomly select 1,000 images per category as the testing queries and 500 per category as the training set. All images except those in the query set serve as the retrieval database. 
(\textbf{iii}) \textbf{NUS-WIDE}~\cite{nuswide} is a large-scale image dataset containing about 270k images from 81 categories. 
We follow \citet{bihalf} to use the 21 most popular categories for evaluation. 100 images per category are randomly selected as the testing queries while the remaining images form the database and the training set.

\subsubsection{Metrics}
We adopt the typical metric, Mean Average Precision (\textbf{MAP}), from previous works~\cite{ssdh,bihalf,cibhash}. 
It is defined as 
\begin{equation}
    \text{MAP@N}=\frac1{|\calQ|}\sum_{\bmx_q\in\calQ}\xiaok{\frac{\sum_{n=1}^N \text{Prec}_q(n)\cdot\mathbb{I}_{\dak{\bmx_n\in\calR_q}}}{|\calR_q|}},
\end{equation}
where $\calQ$ is test query image set and $n$ is the index of a database image in a returned rank list. $\text{Prec}_q(n)$ is the precision at cut-off $n$ in the rank list \wrt $\bmx_q$. $\calR_q$ is the set of all relevant images \wrt $\bmx_q$. $\mathbb{I}_{\dak{\cdot}}$ is an indicator function. 
We follow previous works to adopt MAP@1000 for CIFAR-10 (I) and (II), MAP@5000 for Flickr25K and NUS-WIDE. 

\subsubsection{Models}
We compare the retrieval performance of \textbf{MeCoQ} with \textbf{24} classic or state-of-the-art unsupervised baselines, including: 
(\textbf{i}) 5 shallow hashing methods: 
\textbf{LSH} \cite{lsh}, 
\textbf{SpeH} \cite{speh}, 
\textbf{SH} \cite{sh}, 
\textbf{SphH} \cite{sphh} and  
\textbf{ITQ} \cite{itq}.
(\textbf{ii}) 2 shallow quantization methods: \textbf{PQ} \cite{pq} and
\textbf{OPQ} \cite{opq}. 
(\textbf{iii}) 15 deep binary hashing methods: \textbf{DeepBit} \cite{deepbit}, 
\textbf{UTH} \cite{uth},   
\textbf{SAH} \cite{sah},
\textbf{SGH} \cite{sgh}, 
\textbf{HashGAN} \cite{hashgan}, 
\textbf{GreedyHash} \cite{greedyhash}, 
\textbf{BinGAN} \cite{bingan}, 
\textbf{BGAN} \cite{bgan}, 
\textbf{SSDH} \cite{ssdh}, 
\textbf{DVB} \cite{dvb}, 
\textbf{DistillHash} \cite{distillhash}, 
\textbf{TBH} \cite{tbh}, 
\textbf{MLS$^3$RDUH} \cite{mls3rduh}, 
\textbf{Bi-half Net} \cite{bihalf} and 
\textbf{CIBHash} \cite{cibhash}.
(\textbf{iv}) 2 deep quantization methods:
\textbf{DBD-MQ} \cite{dbdmq} and
\textbf{DeepQuan} \cite{deepquan}. 
We carefully collect their results from related literature. 
When results about some baselines on a certain benchmark are not available (\eg CIBHash on Flickr25K dataset), we try to run their open-sourced codes (if \emph{available} and \emph{executable}) and report the results.

\subsubsection{Implementation Settings}
We implement MeCoQ with Pytorch~\cite{paszke2019pytorch}. 
We follow the standard evaluation protocol~\cite{cibhash,bihalf} of unsupervised deep hashing to use the VGG16~\cite{vgg}. 
Specifically, for shallow models, we extract 4096-dimensional deep $fc7$ features
as the model input. 
For deep models, we directly use raw image pixels as input and adopt the pre-trained VGG16 ($conv1\sim fc7$) as the backbone network. 
We use the data augmentation scheme in \citet{cibhash} that combines random cropping, horizontal flipping, image graying, and randomly applied color jitter and blur.
The default hyper-parameter settings are as follows. 
(\textbf{i}) We set the batch size as $128$ and the maximum epoch as $50$.
(\textbf{ii}) The queue length (\ie the memory bank size), $N_\calM=384$. 
(\textbf{iii}) The smoothness factor of codeword assignment in Eq.(\ref{equ:alpha_softmax}), $\alpha=10$. 
(\textbf{iv}) The codeword number of each codebook, $K=256$ such that each image is encoded by $B=M\log_2K=8M$ bits (\ie $M$ bytes).
(\textbf{v}) The positive prior, $\rho^+=0.1$ for CIFAR-10 (I and II), $\rho^+=0.15$ for Flickr25K and NUS-WIDE.
(\textbf{vi}) The starting epoch for the memory module are set to $5$ on Flickr25K, $10$ on NUS-WIDE and $15$ on CIFAR-10 (I and II).

\begin{table*}[t]
    \centering
    \caption{Mean Average Precision (MAP, \%) results for different MeCoQ variants with different number of bits on Flickr25K, CIFAR-10 (I and II) and NUS-WIDE datasets. The subscript results are the MAP drops compared with full MeCoQ.}
    \resizebox{\linewidth}{!}{
    \setlength{\tabcolsep}{.25em}{
    \begin{tabular}{lllllllllllllllll}
    \toprule
    Dataset $\rightarrow$ &  & \multicolumn{3}{c}{Flickr25K} &  & \multicolumn{3}{c}{CIFAR-10 (I)} &  & \multicolumn{3}{c}{CIFAR-10 (II)} &  & \multicolumn{3}{c}{NUS-WIDE} \\ \cmidrule(l){1-1} \cmidrule(l){3-5} \cmidrule(l){7-9} \cmidrule(l){11-13} \cmidrule(l){15-17}  
    Method $\downarrow$ &  & 16 bits & 32 bits & 64 bits &  & 16 bits & 32 bits & 64 bits &  & 16 bits & 32 bits & 64 bits &  & 16 bits & 32 bits & 64 bits \\ \midrule
    MeCoQ &  & \textbf{81.31} & \textbf{81.71} & \textbf{82.68} &  & \textbf{68.20} & \textbf{69.74} & \textbf{71.06} &  & \textbf{62.88} & \textbf{64.09} & \textbf{65.07} &  & \textbf{80.18} & \textbf{82.16} & \textbf{83.24} \\
    MeCoQ$_{\ \text{w/o debiasing}}$ &  & 80.03$_\text{($\downarrow$\phantom{0}1.28)}$ & 80.19$_\text{($\downarrow$\phantom{0}1.52)}$ & 80.67$_\text{($\downarrow$\phantom{0}2.01)}$ &  & 66.59$_\text{($\downarrow$\phantom{0}1.61)}$ & 67.08$_\text{($\downarrow$\phantom{0}2.66)}$ & 68.77$_\text{($\downarrow$\phantom{0}2.29)}$ &  & 60.51$_\text{($\downarrow$\phantom{0}2.37)}$ & 62.81$_\text{($\downarrow$\phantom{0}1.28)}$ & 62.85$_\text{($\downarrow$\phantom{0}2.22)}$ &  & 79.64$_\text{($\downarrow$\phantom{0}0.54)}$ & 80.68$_\text{($\downarrow$\phantom{0}1.48)}$ & 82.30$_\text{($\downarrow$\phantom{0}0.94)}$ \\
    MeCoQ$_\text{ w/o $\Omega_\calC$}$ &  & 54.00$_\text{($\downarrow$27.31)}$ & 57.53$_\text{($\downarrow$24.18)}$ & 58.84$_\text{($\downarrow$23.84)}$ &  & 39.03$_\text{($\downarrow$29.17)}$ & 36.20$_\text{($\downarrow$33.54)}$ & 42.75$_\text{($\downarrow$28.31)}$ &  & 32.54$_\text{($\downarrow$30.34)}$ & 35.67$_\text{($\downarrow$28.42)}$ & 36.40$_\text{($\downarrow$28.67)}$ &  & 51.33$_\text{($\downarrow$28.85)}$ & 57.29$_\text{($\downarrow$24.87)}$ & 61.01$_\text{($\downarrow$22.23)}$ \\
    MeCoQ$_\text{ w/o $\calM$}$ &  & 78.77$_\text{($\downarrow$\phantom{0}2.54)}$ & 79.50$_\text{($\downarrow$\phantom{0}2.21)}$ & 80.76$_\text{($\downarrow$\phantom{0}1.92)}$ &  & 64.82$_\text{($\downarrow$\phantom{0}3.38)}$ & 67.64$_\text{($\downarrow$\phantom{0}2.10)}$ & 69.36$_\text{($\downarrow$\phantom{0}1.70)}$ &  & 59.78$_\text{($\downarrow$\phantom{0}3.10)}$ & 61.22$_\text{($\downarrow$\phantom{0}2.87)}$ & 63.80$_\text{($\downarrow$\phantom{0}1.27)}$ &  & 76.70$_\text{($\downarrow$\phantom{0}3.48)}$ & 79.13$_\text{($\downarrow$\phantom{0}3.03)}$ & 81.43$_\text{($\downarrow$\phantom{0}1.81)}$ \\
    MeCoQ$_\text{ feature $\calM$}$ &  & 78.39$_\text{($\downarrow$\phantom{0}2.92)}$ & 79.61$_\text{($\downarrow$\phantom{0}2.10)}$ & 80.13$_\text{($\downarrow$\phantom{0}2.55)}$ &  & 65.39$_\text{($\downarrow$\phantom{0}2.81)}$ & 67.62$_\text{($\downarrow$\phantom{0}2.12)}$ & 70.57$_\text{($\downarrow$\phantom{0}0.49)}$ &  & 62.55$_\text{($\downarrow$\phantom{0}0.33)}$ & 63.32$_\text{($\downarrow$\phantom{0}0.77)}$ & 64.84$_\text{($\downarrow$\phantom{0}0.23)}$ &  & 77.71$_\text{($\downarrow$\phantom{0}2.47)}$ & 79.90$_\text{($\downarrow$\phantom{0}2.26)}$ & 81.64$_\text{($\downarrow$\phantom{0}1.60)}$ \\
    MeCoQ$_\text{ hard code $\calM$}$ &  & 60.77$_\text{($\downarrow$20.54)}$ & 63.10$_\text{($\downarrow$18.61)}$ & 68.49$_\text{($\downarrow$14.19)}$ &  & 33.95$_\text{($\downarrow$34.25)}$ & 40.65$_\text{($\downarrow$29.09)}$ & 46.76$_\text{($\downarrow$24.30)}$ &  & 34.40$_\text{($\downarrow$28.48)}$ & 37.55$_\text{($\downarrow$26.54)}$ & 44.90$_\text{($\downarrow$20.17)}$ &  & 69.46$_\text{($\downarrow$10.72)}$ & 72.23$_\text{($\downarrow$\phantom{0}9.93)}$ & 74.95$_\text{($\downarrow$\phantom{0}8.29)}$ \\
    MeCoQ$_\text{ w/o delaying $\calM$}$ &  & 79.02$_\text{($\downarrow$\phantom{0}2.29)}$ & 78.44$_\text{($\downarrow$\phantom{0}3.27)}$ & 78.17$_\text{($\downarrow$\phantom{0}4.51)}$ &  & 66.76$_\text{($\downarrow$\phantom{0}1.44)}$ & 67.82$_\text{($\downarrow$\phantom{0}1.92)}$ & 69.04$_\text{($\downarrow$\phantom{0}2.02)}$ &  & 59.79$_\text{($\downarrow$\phantom{0}3.09)}$ & 61.42$_\text{($\downarrow$\phantom{0}2.67)}$ & 62.60$_\text{($\downarrow$\phantom{0}2.47)}$ &  & 79.72$_\text{($\downarrow$\phantom{0}0.46)}$ & 81.39$_\text{($\downarrow$\phantom{0}0.77)}$ & 82.67$_\text{($\downarrow$\phantom{0}0.57)}$ \\ \bottomrule
    \end{tabular}
    }}
    \label{tab:map2}
\end{table*}


\subsection{Results and Analysis}
\subsubsection{Comparison with Existing Methods}
The MAP results in Tables~\ref{tab:map1} show that MeCoQ substantially outperforms all the compared methods. 
Specifically, compared with CIBHash, a latest and strong baseline, MeCoQ achieves average MAP increases of \textbf{3.52}, \textbf{6.92}, \textbf{2.24} and \textbf{1.46} on Flickr25K, CIFAR-10 (I), (II) and NUS-WIDE datasets, respectively. 
Besides, we can get two findings from the MAP results. 
(\textbf{i}) Deep methods do not always outperform shallow methods with CNN features. 
For instance, DeepBit and UTH do not outperform PQ and OPQ with CNN features on NUS-WIDE. 
It implies that without label supervision, some deep hashing methods are less effective to take good advantage of pre-trained CNNs. 
(\textbf{ii}) Contrastive Learning (CL) is effective to learn deep hashing models.
The two CL-based methods in Table~\ref{tab:map1}, CIBHash and MeCoQ, perform best on all datasets.
Moreover, with debiasing mechanism and code memory, MeCoQ shows notable improvements over CIBHash. 


\begin{figure}[t]
  \centering
  \includegraphics[width=0.9\columnwidth]{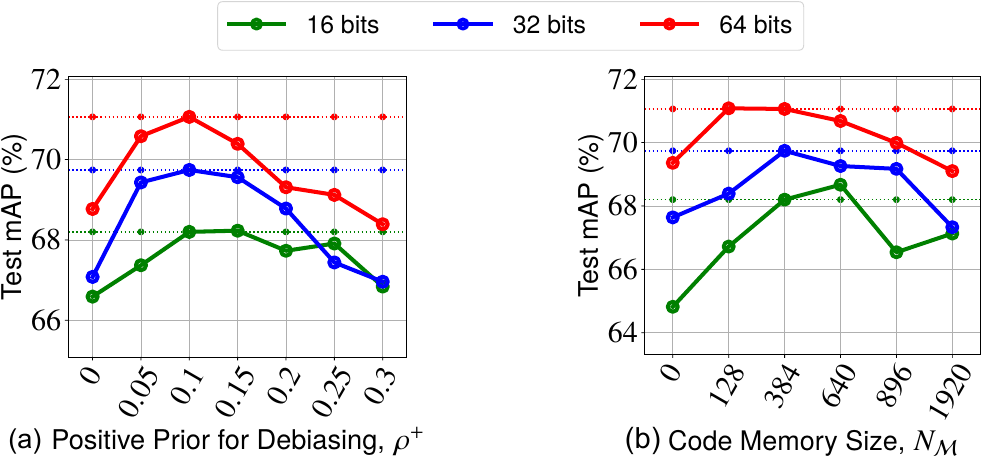}
  \caption{Sensitivities of $\rho^+$ and $N_\calM$ on CIFAR-10 (I). The dotted lines indicate the MAP results of default settings.}
\label{fig:hp}
\end{figure}

\subsubsection{Component Analysis}
We set 6 MeCoQ variants to analyse the contributions of components: 
(\textbf{i}) MeCoQ$_{\ \text{w/o debiasing}}$ removes debiasing mechanism by setting $\rho^+=0$ in Eq.(\ref{equ:inbatch}) and (\ref{equ:memory});
(\textbf{ii}) MeCoQ$_\text{ w/o $\Omega_\calC$}$ removes codeword regularization, $\Omega_\calC$;
(\textbf{iii}) MeCoQ$_\text{ w/o $\calM$}$ removes the code memory, $\calM$;
(\textbf{iv}) MeCoQ$_\text{ feature $\calM$}$ replaces soft code memory by feature memory;
(\textbf{v}) MeCoQ$_\text{ hard code $\calM$}$ replaces soft code memory by hard code memory;
(\textbf{vi}) MeCoQ$_\text{ w/o delaying $\calM$}$ begins memory-augmented training at the very start of the learning. We can make the following summaries based on the Table~\ref{tab:map2}.

\textbf{Debiasing improves performance.}
MeCoQ outperforms MeCoQ$_{\ \text{w/o debiasing}}$ by 1.60, 2.19, 1.96 and 0.99 of average MAPs on Flickr25K, CIFAR-10 (I), (II) and NUS-WIDE, which shows that correcting the sampling bias can improve model training. 
Figure~\ref{fig:hp}(a) shows that the optimal $\rho^+$ on CIFAR-10 (I) is about 0.1, which means that we may randomly drop a false-negative sample from the training set with a 10\% probability. 
It is consistent with the property of CIFAR-10 (I) that each category accounts for 10\%.

\textbf{The codeword diversity regularization avoids model degeneration.}
MeCoQ outperforms MeCoQ$_\text{ w/o $\Omega_\calC$}$ by 25.11, 30.34, 29.14 and 25.32 of average MAPs on 4 datasets. It demonstrates the importance of regularization. 

\textbf{Soft code memory is effective and efficient to enhance contrastive learning.}
MeCoQ outperforms MeCoQ$_\text{ w/o $\calM$}$ by 2.22, 2.39, 2.41 and 2.77 of average MAPs on 4 datasets, which verifies the worth of using $\calM$. 
Besides, Figure~\ref{fig:hp}(b) shows that enlarging memory to cache more negatives improves MeCoQ, while the improvement tends to drop as memory size exceeds a certain range. 
The reason is that the low feature drift phenomenon only holds within a limited period. 
We can also learn that fewer bits with larger quantization errors allow a slightly larger memory because the quantization error can partially offset the feature drift. 
Moreover, as shown in Table~\ref{tab:map3}, using code memory is efficient in GPU memory and computation. 
It can achieve better results with much less GPU memory than enlarging batch size. 
The marginal increase of time is caused by similarity computations between training queries and cached negatives. 


\textbf{Soft code memory is better than feature memory and hard code memory.}
MeCoQ outperforms MeCoQ$_\text{ feature $\calM$}$ by 2.52, 1.81, 0.44 and 2.11 of average MAPs on 4 datasets, because the soft code memory has lower feature drift than feature memory. 
Surprisingly, MeCoQ$_\text{ hard code $\calM$}$ fails. 
It seems that reconstructed features from hard codes become adverse noises rather than valid negatives because the large error of hard quantization leads to an over-large feature drift.

\textbf{It is better to delay the usage of memory in the learning process.}
MeCoQ outperforms MeCoQ$_\text{ w/o delaying $\calM$}$ by 3.36, 1.79, 2.74 and 0.6 of average MAPs on 4 datasets. It implies that using code memory from the very beginning leads to sub-optimal solutions because reusing unstable representations in initial training stage is not recommended.

\begin{table}[t]
    \centering
    \caption{Model profiling results on CIFAR-10 (I) dataset, including the number of negative samples (`\#Neg.') per training query, average training time per epoch in seconds (`Time / Ep. / sec'), GPU memory demand in megabytes (`GPU Mem. / MB') and MAP (\%) for 32 bits, under different batch size and memory size settings. 
    $N$ and $N_\calM$ denote the batch size and memory size respectively.
    We do these experiments on a single NVIDIA GeForce GTX 1080 Ti (11GB) and Intel® Xeon® CPU E5-2650 v4 @ 2.20GHz (48 cores).
    }
    \resizebox{\linewidth}{!}{
    \setlength{\tabcolsep}{.25em}{
    \begin{tabular}{llclclclc}
    \toprule
    Method && \#Neg. && Time / Ep. / sec && GPU Mem. / MB && MAP(\%)@32 bits \\
    \midrule
    $N$=128, w/o $\calM$ && 128 && 528\phantom{$_\text{($\uparrow$00)}$} && \phantom{0}5925\phantom{$_\text{($\uparrow$5002)}$} && 67.64\phantom{$_\text{($\uparrow$0.00)}$} \\
    \midrule
    $N$=256, w/o $\calM$ && 256 && 550$_\text{($\uparrow$22)}$ && 10927$_\text{($\uparrow$5002)}$ && 68.66$_\text{($\uparrow$1.02)}$ \\
    $N$=128, $N_\calM$=128 && 256 && 536$_\text{($\uparrow$8)\phantom{0}}$ && \phantom{0}5927$_\text{($\uparrow$2)\phantom{000}}$ && 68.39$_\text{($\uparrow$0.75)}$ \\
    $N$=128, $N_\calM$=384 && 512 && 547$_\text{($\uparrow$19)}$ && \phantom{0}5933$_\text{($\uparrow$8)\phantom{000}}$ && 69.74$_\text{($\uparrow$2.10)}$ \\
    $N$=128, $N_\calM$=896 && 1024 && 563$_\text{($\uparrow$35)}$ && \phantom{0}5947$_\text{($\uparrow$22)\phantom{00}}$ && 69.17$_\text{($\uparrow$1.53)}$ \\
    \bottomrule
    \end{tabular}
    }}
    \label{tab:map3}
\end{table}

\section{Conclusions}
In this paper, we propose \textbf{Co}ntrastive \textbf{Q}uantization with Code \textbf{Me}mory (\textbf{MeCoQ}) for unsupervised deep quantization.
Different from existing reconstruction-based unsupervised deep hashing methods, MeCoQ learns quantization by contrastive learning. 
To avoid model degeneration when optimizing MeCoQ, we introduce a codeword diversity regularization. 
We further improve the memory-based contrastive learning by designing a novel quantization code memory, which shows lower feature drift than existing feature memories without using momentum encoder. 
Extensive experiments show the superiority of MeCoQ over the state-of-the-art methods. 
More importantly, MeCoQ sheds light on a promising future direction to unsupervised deep hashing. Potentials of contrastive learning remain to be explored.

\section{Acknowledgments}
This work is supported in part by the National Key Research and Development Program of China under Grant 2018YFB1800204, the National Natural Science Foundation of China under Grant 61771273 and 62171248, the R\&D Program of Shenzhen under Grant JCYJ20180508152204044, and the PCNL KEY project (PCL2021A07).

\bibliography{aaai22}


\end{document}